\pdfoutput=1
\documentclass[conf]{new-aiaa}
\usepackage{booktabs}
\usepackage{graphicx}
\usepackage{natbib}
\usepackage{xcolor}
\usepackage{amsmath}
\usepackage{bm}
\usepackage{float}
\usepackage{subcaption}

\title{Sequential Path Planning with a Multi-Vehicle Team  \\ Through Uncertain Risk Regions}
\author{Artur Wolek\footnote{Assistant Professor, Department of Mechanical Engineering and Engineering Science, AIAA Senior Member.}, Dipankar Maity\footnote{Associate   Professor, Department of Electrical and Computer Engineering.}}
\affil{
University of North Carolina at Charlotte, Charlotte, North Carolina, 28223}
\author{
Alexander Von Moll\footnote{Electronics Engineer, Control Science Center, AIAA Member.}, Satyanarayana Manyam\footnote{DCS Corp Contractor.}, Thomas Chapman\footnote{Control Science Center.}, David Casbeer\footnote{Technical Area Lead for UAV Cooperative and Intelligent Control, AIAA Associate Fellow.}, Isaac Weintraub\footnote{Electronics Engineer, Control Science Center, AIAA Associate Fellow.}
}
\affil{
Air Force Research Laboratory, Wright-Patterson AFB, OH 45433, USA}

\begin{document}
\maketitle

\begin{abstract}
This paper considers the problem of planning a path for a large multi-vehicle team (i.e., swarm) of low-cost vehicles to reach a goal location in the presence of uncertain risk regions. The risk regions have a known circular radius but their position in a bounded, two-dimensional,  rectangular environment is not known to the swarm. Swarm agents are sequentially released from the start to the goal but when they enter a risk region they are disabled with some probability. The path of the previous swarm agent is then used to update an evolving model of the environment through recursive Bayesian estimation. The aim of the swarm is to reach the goal with the fewest number of disabled swarm agents. An algorithm is proposed to meet this objective while considering risk in path planning and balancing exploration versus  exploitation to identify a feasible path. Subsequently, a second algorithm aims to optimize the feasible path by reducing its length. The paper presents an illustrative example of the approach.
\end{abstract}
\let\thefootnote\relax\footnotetext{Distribution Statement A. Approved for public release: distribution is unlimited. AFRL-2026-2612. Cleared 28 May 2026.}

\section{Introduction}

\lettrine{T}here are many contexts in which a mobile robot (e.g., ground vehicle or unmanned aircraft) may enter hazardous regions that pose a danger to the robot itself. Examples include  flooded regions, wildfires \cite{aggarwal2022risk}, disaster zones, regions with interfering agents, or contaminated areas \cite{rosas2020autonomous}. The majority of prior work focusing on navigating around risk regions has generally assumed the position of the risk regions to be known. For example, authors have studied  path planning to a goal while minimizing detection from radar sites using the A* algorithm \citep{zhang2020novel, zhang2023efficient} and in 3-D using particle swarm optimization, grey wolf optimization, and genetic algorithms \citep{xu2020optimized}. Other approaches that account for the dynamic nature of a single risk region \citep{weintraub2022optimal} and two risk regions \citep{dillon2023optimal} are based on the calculus of variations and direct optimal control.  More recently, some authors have started to consider probabilistic and uncertain risk regions. For example, \cite{stagg2026inferring} considers a geometric-reachable-region model  to infer parameters of uncertain risk regions using sacrificial agents in a continuous space with rigorous bounds. Also, \cite{stagg2026engagement} investigates determining feasible pursuer launch regions based on knowledge of interception inside risk regions. Past models have assumed risk regions with a cardioid shape as a surrogate geometry since its shape is representative  \citep{herrmann1996air, yan2020reachability}. Tuning the shape to match an existing risk model is out of the scope of this work. Further discussion of risk region models for aircraft modeled using simple motion (i.e., no minimum turning radius) can be found in \citep{von2023basic}.

In this problem formulation, we consider the path of the swarm agents prior to being disabled as a key source of information used to update the model of the environment. This line of reasoning with low-cost agents is closely related to the methods proposed by Otte et al. \cite{mcguire2024valuing, otte2021path}. The work in \cite{mcguire2024valuing} quantifies how the loss of low-cost agents along their paths can be used to infer hazard locations and guide information-gathering in communication-restricted environments. Similarly, \cite{otte2021path} considers that an agent was lost somewhere along a path, but that there is no knowledge of where along the path that event occurred. This paper builds on our prior work by solving an uncertain version of the problem investigated in \cite{wolek2024sampling} where the risk regions had a known position and only a single flight vehicle was considered planning a path. Here, a team of swarm agents is tasked with navigating around an unknown set of risk regions to reach a goal. The risk regions are modeled as having a known circular geometry centered around a {generator} point that is the origin of each risk region. Swarm agents are sequentially released from the starting point to the goal, and if they enter a risk region there is a probability that they will be lost (i.e., fail or become immobilized). The location of where the agent is lost can be used to infer the locations of the generator points and thereby the geometry of the risk regions. The objective for the swarm is to reach the goal with the fewest number of disabled swarm agents.

\subsection{Contributions}
The contributions of this paper are: (1) a recursive Bayesian estimation framework that updates cell-level generator probabilities using both survival and disabling events along executed swarm-agent paths, enabling principled inference of uncertain risk region locations; and (2)  a risk-aware swarm path-planning algorithm that couples the evolving generator-probability field with an exploration--exploitation policy to identify a safe passage through multiple uncertain risk regions with minimal attrition and minimum path length. The performance of the algorithm is characterized through a numerical example that determines how many swarm agents are needed to successfully reach a target area.

\subsection{Organization}
The remainder of the paper is organized as follows. Section~\ref{sec:problem_formulation} formulates the problem to be solved by describing the assumptions regarding the environment, risk regions, agent attrition, and stating the path planning objective. Section~\ref{sec:methadology} presents the proposed methodology. The estimation approach is presented and then the risk-based path planner is proposed. Section~\ref{sec:results} provides an illustrative example. Section~\ref{sec:conclusion} concludes the paper.

\section{Problem Formulation}
\label{sec:problem_formulation}
\begin{figure}[h!]
    \centering
    \includegraphics[width=0.45\linewidth]{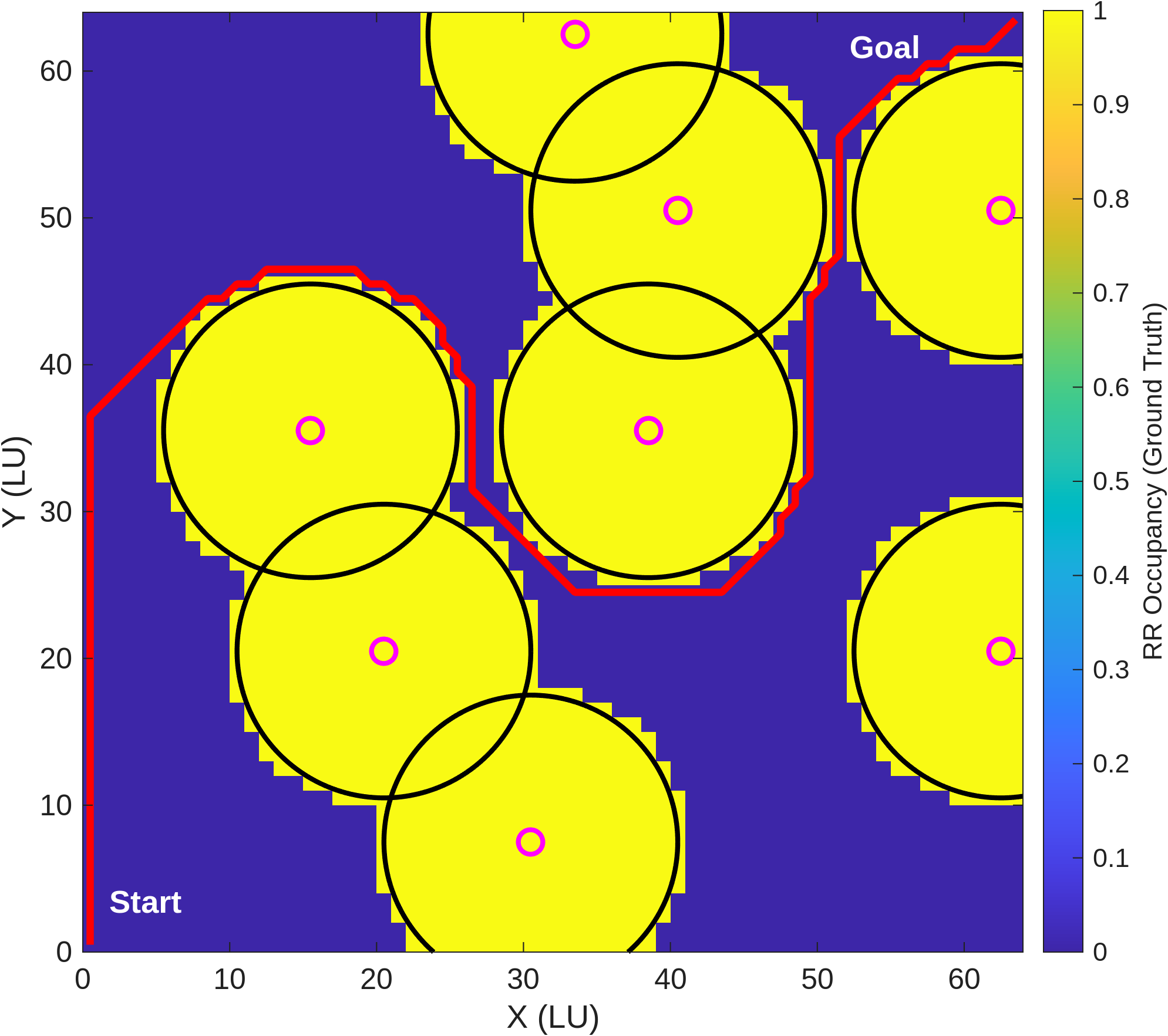}
    \caption{The problem of this paper is to find a path from the start to the goal while navigating around risk regions that are initially unknown. The image above illustrates a shortest path in a discretized environment if the risk regions were known exactly.}
    \label{fig:setup}
\end{figure}

Consider a large swarm of $N$ low-cost agents operating in a contested two-dimensional planar environment in the presence of $M$ risk regions. The operating domain is considered to be a square  set, $Z = [0, H]^2$, with the swarm agents beginning initially at the origin (i.e., lower left corner). The  objective of the swarm is to reach the goal located in the upper right corner (see Fig.~\ref{fig:setup}). The risk regions are modeled as discs of radius $R_m$ centered on positions $(x_m, y_m) \in Z$ for $m=1,\ldots, M$. The swarm agents have no prior knowledge of the risk-region positions and upon entering a risk region
\[
C_m := \{ (x,y) \in \mathbb{R}^2~:~\lVert( x-x_m, y-y_m)\rVert \leq R_m \}
\]
the swarm agents {may} be disabled. Whether a risk region disables a swarm agent is an uncertain outcome. The particular model describing this interaction is described in the next section.
The aim of the swarm is to reach the goal location quickly and with minimal number of total disabled agents. The path that each swarm agent takes prior to being disabled is known to a centralized processor that uses the information to recursively update an estimate of possible risk regions and guide subsequent swarm agents. While the objective of the swarm is not to map the locations of the risk regions throughout all of $Z$, the swarm agents must gain sufficient understanding of risk regions to identify a safe passage. As a secondary objective, shorter low-risk paths are preferred.

\subsection{Environment Model}
The environment is discretized into an $S$-by-$S$ grid of square cells with length and width $\Delta$ (i.e., $H/S = \Delta$). The first cell (corresponding to the start location) is labeled with the index 1, and subsequent cells are labeled in the order increasing to the right, until the end of the row (cell $S$) and the cell $S+1$ appears above cell 1 on the subsequent row. The labels continue in this manner until the last cell at the goal location is reached, which is indexed $S^2$.
Let $(x_i, y_i)$ denote the center location of the $i$th cell. The corresponding cell area is
\[
A_i := \left[x_i - \frac{\Delta}{2},x_i + \frac{\Delta}{2} \right ] \times
\left[y_i - \frac{\Delta}{2},y_i + \frac{\Delta}{2} \right ] \;.
\]
The swarm agents move over a directed graph $\mathcal{G}= (V,E)$ with vertices $V = \{1, \ldots, S^2\}$ corresponding to each cell center and edges $E \subset V \times V$. Each edge $e \in E$ is a pair of vertices, i.e., $e = (v_i, v_j)$ that represents a transition from vertex $v_i \in V$ to vertex $v_j \in V$. The swarm agents begin at vertex $v_1$ in the first cell and aim to reach the vertex $v_{S^2}$ in the last cell of the discretized environment. Edges exist between pairs of vertices that are adjacent to one another in a Moore neighborhood.
The edge motions correspond to motion along the cardinal directions $\{ {\rm N, E, S, W} \}$ with edge lengths $\Delta$, or along the ordinal directions  $\{ {\rm SE, SW, NW, NE}\}$  with edge lengths $\sqrt{2} \Delta$ (except at the boundaries where motion leaving the domain $Z$ is  not admissible). The mapping $\texttt{Length}~:E \to \{ \Delta, \sqrt{2} \Delta \}$ maps each edge to the corresponding length.
\subsection{Risk-Region Model}
Each cell has a type $O_i \in \{0, 1\}$, and either $O_i = 1$ if the cell is a {risk-region cell} or $O_i = 0$ if the cell is a {free cell}. Risk-region cells are within the range of one or more generators and by entering the cell an agent may be disabled. Free cells are other cells in the environment where there is no risk of being disabled. In Fig.~\ref{fig:setup} the risk-region cells are shown in yellow, whereas the free cells are shown in blue.

The type of the cell depends on its proximity to nearby risk regions. Cell $i$ is considered to be a risk-region cell if any part of the cell overlaps with one or more risk regions, i.e., $O_i = 1$ if
\[
A_i \cap \bigcup_{m=1}^M C_m \neq \emptyset
\]
and otherwise the cell is a free cell, $O_i = 0$. Conceptually, the cells where $O_i = 1$ can be viewed as stochastic obstacles (unknown to the swarm) to be avoided in path planning. The obstacles are stochastic because it is possible for an agent to enter the cell and later exit without being disabled (as will be described in the following). The starting node and goal node are assumed to be safe $O_1 = 0$ and $O_{S^2} = 0$.

A path starting from the start location is an ordered list of $Q$ vertices:
\[
\pi := \{u_1, \ldots, u_Q \} \;,
\]
with $u_1 = v_1$ and $u_Q \in V$. If the path terminates at the goal, then $u_Q= v_{S^2}$. Each ordered pair of vertices in $\pi$ is an admissible motion (i.e., $(u_i, u_{i+1}) \in E$ for all $i =1, \ldots, Q-1$). The graph contains no self-loops so the agent must continue to move  along edges. We view each edge transition as occurring sequentially in time (approximately equal to one time step).
The path is guaranteed to be safe if $O_{u_i} = 0$ for all $u_i \in \pi$ and we denote this condition as $\texttt{Safe}(\pi) =1$, and otherwise $\texttt{Safe}(\pi) =0$ for unsafe paths.  If a swarm agent visits a vertex $v_i$ for which $O_i = 1$ then the generator for that risk cell disables the swarm agent with probability $p_f$.  Each visit to a risk-region cell is considered an independent Bernoulli trial with probability $p_f$ (constant across all risk regions). If multiple risk regions overlap a cell then there are multiple Bernoulli trials to be considered and any of these succeeding results in the agent being disabled. The Bernoulli trials that determine whether an agent is disabled occur only once upon entering a cell from a unique cell along an edge.

For the remainder of this work, we restrict the risk region  centers to coincide exactly with cell center locations (i.e., every risk-region center $(x_m, y_m)$ for $m = 1, \ldots, M$ is precisely at the center of some cell whose index is denoted $q(m) ~:~\{1,\ldots,M\} \to \{1, \ldots, S^2\}$). Furthermore, we assume that the risk-region radii are all uniform (i.e., $R_m = R$ for $m=1,\ldots,M$) and both the risk-region radius $R$ and probability of disablement $p_f$ are known to the swarm agents.
\subsection{Problem Statement}
\label{sec:prob_statement}
For the problem to be well posed, we assume that there exists a safe path  that joins start and goal nodes. An algorithm (to be designed) generates a sequence of paths $\hat \pi_1, \ldots, \hat \pi_N$ that are each sequentially executed by one swarm agent per path. That is, a swarm agent plans a path $\hat \pi_n$ but only executes a subset of the path $\pi_n \subseteq \hat \pi_n$ if it is disabled prior to reaching the goal. The path $\pi_n$ is available to the $(n+1)$th agent to consider when planning the next path $\pi_{n+1}$. The objective is to identify a safe path with the fewest number of agents.
\begin{align*}
\min_{N \in \mathbb{Z}^+} N ~\text{such that there exists}~ n \in \{1, \ldots, N\}~\text{with}~ \texttt{Safe}(\pi_n) = 1 \;.
\end{align*}
Because the swarm does not have perfect knowledge of the environment, the condition $\texttt{Safe}(\pi_n) = 1$ cannot be evaluated with certainty. Instead, we will replace this deterministic constraint with a chance constraint after the notion of path {feasibility} is introduced.
\section{Methodology}
\label{sec:methadology}
This section describes the environment estimation algorithm and an approach to construct minimum risk paths to solve the above problem.
\subsection{Generator Estimation via Recursive Bayesian Estimation}
Recall that the cells in the environment are indexed as $ k \in \{1, \ldots, S^2\}$.
For each cell $k$ in the environment we define three random Bernoulli variables:
\begin{enumerate}
    \item $G_k \in \{0, 1\}$ describes  whether cell $k$ is the center location of a risk region ($G_k = 1$), and we call such cells {generators},
    \item $F_{kj} \in \{0, 1\}$ describes the event of a generator at cell $k$ disabling a swarm agent in cell $j$ during a given time step $(F_{kj} = 1)$,
    \item $D_k \in \{0, 1\}$ describes the aggregated outcome of all risk regions, and whether a swarm agent that spends one time step in cell $k$ is disabled $(D_k = 1)$.
\end{enumerate}
The random variables $G_k$ are independent across all cells, and the agent-generator interactions are independent across time steps. Once a swarm agent is deployed it is assumed to move to an adjacent cell at each time step (i.e., consecutive elements of $\hat \pi_n$ are unique and considered to be consecutive time steps). Let $\rho(j)$ denote the set of vertices that are within a distance $R$ of cell $j$, i.e.,
\begin{align*}
\rho(j) := \{ i \in \{1, \ldots, S^2\} ~:~ \texttt{dist}((x_j, y_j), A_i) \leq R \} \;,
\end{align*}
where
\begin{equation*}
    \texttt{dist}((x_j, y_j), A_i) := \sqrt{{\rm max}(|x_i - x_j| - \Delta/2,0)^2 + {\rm max}(|y_i - y_j| - \Delta/2,0)^2} \;
\end{equation*}
is a distance function that  ensures that even if a small portion of the rectangular region of cell $i$, covering the area $A_i$ with center $(x_i,y_i)$, overlaps with the disk of a generator $j$ it is included in the set $\rho(j)$. The set $\rho(j)$ is the set of cells that pose a risk of disabling the agent located at cell $j$.

Consider a different set of cells centered around a possible generator cell $a$ (rather than centered around the agent).
The probability that cell $a$ disables a swarm agent in cell $b \in \rho(a)$ is conditioned on the cell $a$ being a generator as follows:
\begin{equation}
\begin{split}
    P( F_{ab}=1 | G_a=1 ) &= p_f \;, \\
    P( F_{ab}=1  | G_a=0  ) &= 0 \;, \\
    P( F_{ab}=0  | G_a=1 ) &= 1- p_f \;, \\
    P( F_{ab}=0  | G_a=0  ) &= 1 \;,
    \end{split} \label{eq:pfg_rhoi}
\end{equation}
where $0 < p_f < 1$ is a known probability and, if $b \notin \rho(a)$, then a generator at cell $a$ is too far away to affect cell $b$:
\begin{equation}
\begin{split}
    P( F_{ab}=1 | G_a=1 ) &= 0 \;, \\
    P( F_{ab}=1  | G_a=0  ) &= 0 \;, \\
    P( F_{ab}=0  | G_a=1 ) &= 1 \;, \\
    P( F_{ab}=0  | G_a=0  ) &= 1 \;.
    \end{split} \notag
\end{equation}

Now, returning to the agent-centric perspective, suppose that an agent enters cell $j$ and, for one time step, it remains unharmed $(D_j = 0)$. This increases the likelihood that the cells in $\rho(j)$ contain no generators. The agent survives if and only if all generators in $\rho(j)$ do not disable (i.e., the joint probability that all $F_{ij}$ are zero):
\begin{align}
P(D_j=0) = \underbrace{\prod_{i\in \rho(j)} P(F_{ij}=0)}_{\text{prob. that no generators in $\rho(j)$ successfully disable agent in $j$}} \;.
\label{eq:P_Di}
\end{align}
For convenience, define $g_i := P(G_i=1)$, thus  $P(G_i=0) = 1-g_i$.
Compute the marginal likelihood that $F_{ij} = 1$ if $j \in \rho(i)$:
\begin{align*}
P(F_{ij} = 1) &= P(F_{ij}=1|G_i=0)P(G_i=0) + P(F_{ij}=1|G_i=1)P(G_i=1)  \\
&= p_fg_i
\end{align*}
and the marginal likelihoods that $F_{ij} = 0$ if $j \in \rho(i)$:
\begin{equation}
\begin{split}
P(F_{ij} = 0) &= P(F_{ij}=0|G_i=0)P(G_i=0) + P(F_{ij}=0|G_i=1)P(G_i=1)  \\
&= 1(1-g_i) + (1-p_f)g_i  \\
&= 1 - g_i + g_i - p_f g_i  \\
&= 1 - p_f g_i \;.
\end{split}   \label{eq:F_j=0}
\end{equation}
If $j \notin \rho(i)$ then $P(F_{ij} = 1)  = 0$ and $P(F_{ij} = 0)=1$. Using \eqref{eq:F_j=0}, the likelihood \eqref{eq:P_Di} becomes
\begin{equation}
P(D_j=0) =   \prod_{k \in \rho(j)} (1-p_fg_k)\;.
\label{eq:prob_Di=0}
\end{equation}
The outcome of an agent surviving at cell $j$ ($D_j = 0$) can be conditioned on knowledge of one cell $i$ being a generator (while others remain random). This is achieved by decomposing the product \eqref{eq:P_Di} into a conditional term:
\begin{align}
P(D_j=0|G_i =1) &= \prod_{k\in \rho(j)} P(F_{kj}=0|G_i =1) \notag \\
&= P(F_{ij}=0|G_i=1) \prod_{k \in \rho(j), k\neq i} P(F_{kj}=0) \label{eq:pdjgi1} \;,
\end{align}
since the generators are independent and thus $P(F_{kj}=0|G_i =1) = P(F_{kj}=0)$ for all $k \neq i$.
Furthermore,  \eqref{eq:pdjgi1} can be expanded using the law of total probability and  simplified with the definition of $g_k$ and \eqref{eq:pfg_rhoi}:
\begin{align}
P(D_j=0|G_i=1) &= P(F_{ij}=0|G_i=1) \prod_{k \in \rho(j), k\neq i} \left[  P(F_{kj}=0|G_k=0)P(G_k=0) + P(F_{kj}=0|G_k=1)P(G_k=1) \right] \notag \\
&= (1 - p_f) \prod_{k \in \rho(j), k\neq i} [1(1-g_k) + (1-p_f)g_k] \notag \\
&= (1 - p_f) \prod_{k \in \rho(j), k\neq i} (1-p_fg_k) \;. \label{eq:p(di=0|gj=0)}
\end{align}
If the agent survives at cell $j$ for one time step we treat this evidence, that $D_j = 0$, as a measurement and use Bayes' rule to update the probability of nearby generators. The update for all generator cells in $\rho(j)$ using Bayes' rule with \eqref{eq:p(di=0|gj=0)}, \eqref{eq:prob_Di=0} is
\begin{align}
       P(G_i=1 | D_j=0 ) &= \frac{P(D_j = 0|G_i=1) P(G_i=1)}{P(D_j=0)} \notag \\
       g_i' &= \frac{\left[  (1 - p_f) \prod_{k \in \rho(j), k\neq i} (1-p_fg_k) \right]g_i}{ \prod_{k \in \rho(j)} (1-p_fg_k) } \notag \\
       &= \frac{   (1 - p_f) g_i}{ 1-p_fg_i }  \label{eq:prob_gj=1|di=0} \;,
\end{align}
where the product terms canceled, $g_i'$ denotes the updated likelihood of the generator (posterior), $P(G_i=1)$ is the prior, and $P(D_j=0)$ is the marginal likelihood of survival. Equation \eqref{eq:prob_gj=1|di=0} thus provides a means to update the generator probabilities given evidence that an agent survived one time step in cell $j$.
Now suppose that an agent enters cell $j$ and is disabled. This event is equal to the probability that at least one generator disables at cell $j$, which is given by the complement of \eqref{eq:prob_Di=0}:
\begin{align}
P(D_j=1) &=  1 - \underbrace{\prod_{i\in \rho(j)} P(F_{ij}=0)}_{\text{prob. that no generators disable}} \notag \\
&= \underbrace{1 - \prod_{i \in \rho(j)}(1-p_fg_i)}_{\text{prob. that at least one generator disables}} \label{eq:prob_di=1} \;.
\end{align}
Now consider that the probability of the event in \eqref{eq:prob_di=1} is conditioned on a generator being present at cell $i$ (assuming $i \in \rho(j)$). Following a conditional form of \eqref{eq:prob_di=1} and using \eqref{eq:pfg_rhoi} and \eqref{eq:F_j=0}:
\begin{align}
P(D_j=1|G_i=1)
&=  1 - \underbrace{\prod_{k\in \rho(j)} P(F_{kj}=0|G_i=1)}_{\text{prob. that no generators disable given knowledge of $G_i$}} \notag \\
&= 1 - P(F_{ij}=0|G_i = 1) \prod_{k \in \rho(j), k\neq i} P(F_{kj}=0|G_i = 1) \notag \\
&= 1 -  (1-p_f) \prod_{k \in \rho(j), k\neq i} P(F_{kj}=0) \notag \\
&= 1 -  (1-p_f) \prod_{k \in \rho(j), k\neq i} (1-p_f g_k) \label{eq:prob_di=1|gj=1}
\;.
\end{align}
As before, we may use Bayes' rule to update the probability of generators in cells $\rho(j)$. Given that an agent is disabled in cell $j$ the probability of generators is updated with  \eqref{eq:prob_di=1} and \eqref{eq:prob_di=1|gj=1} as:
\begin{align}
       P(G_i=1 | D_j=1 ) &= \frac{P(D_j = 1|G_i=1) P(G_i=1)}{P(D_j=1)} \notag \\
       g_i' &= \frac{\left(  1 -  (1-p_f) \prod_{k \in \rho(j), k\neq i} (1-p_f g_k) \right) g_i}{1 - \prod_{k \in \rho(j)}(1-p_fg_k)}
       \label{eq:prob_gj=1|di=1} \;.
\end{align}

Thus, we have established a means to update the probability of generators, via \eqref{eq:prob_gj=1|di=0} and \eqref{eq:prob_gj=1|di=1}, in the environment by considering the path of swarm agents and the aggregated outcomes of where they survive and where they are disabled along the path. The equations must be applied sequentially after each step along the path.

We note here that our recursive updates derived above occur cell-wise rather than over the full joint distribution of all cells. Each of the $S^2$ cells can have one of two states, so an exact Bayesian inference would maintain probability over all possible binary combinations of these cells (i.e., $2^{(S^2)}$ possible cell realizations). Survival events that update cells via \eqref{eq:prob_gj=1|di=0} correctly treat all updated cells as independent (i.e., $g_i'$ only depends on $g_i$); however, disablement events \eqref{eq:prob_gj=1|di=1} depend not only on cell $i$ but also on all cells in $\rho(j)$. Thus the disablement update introduces correlation which, in an exact Bayesian framework, should be represented over the full joint state space across all cells. However, for tractability, we ignore this correlation (i.e., we assume a mean-field approximation) and maintain the probability over each cell  independently.

\subsection{Spatial Risk Modeling}
The estimation approach described above provides an estimate of probable locations of generators in the environment. However, the spatial map of possible generator positions is not directly useful for avoiding risk regions since generators act at a distance $R$. To represent the level of risk in the environment we introduce a fourth Bernoulli variable in each cell: $W_j \in \{0, 1\}$. $W_j$ describes whether cell $j$ is within the disk of at least one risk-region generator ($W_j = 1$). The probability that at least one cell in $\rho(j)$ is a generator is a result of the complement:
\begin{align}
P(W_j=1) &=  1 - \underbrace{\prod_{i\in \rho(j)} (1 - P(G_i=1))}_{\text{prob. that no cells in $\rho(j)$ are generators}}  \notag \\
&= \underbrace{1-\prod_{i\in \rho(j)} (1- g_i)}_{\text{prob. that at least one cell in $\rho(j)$ is a generator}}  \;.\label{eq:RR_prob}
\end{align}
For convenience, define $w_j := P(W_j = 1)$. Computing the above quantity for each cell in the environment provides a spatial risk map that can be used for path planning.

\subsection{Path Planning Algorithms}
This section outlines two path planning algorithms that can be used in two consecutive phases. The first algorithm identifies a {feasible} path (satisfying a chance constraint), and the second algorithm provides a method to reduce the length of this feasible path. As discussed in Sec.~\ref{sec:prob_statement}, the objective is to find a feasible path with the fewest number of agents and a secondary objective is for this path to be as short as possible. To guide the path planning algorithm using the spatial risk map ($w_i$) we first introduce three distinct types of paths that vary in how they quantify path cost. Next, we quantify path feasibility, and lastly, we present the two algorithms for each phase.
\subsubsection{Path Types I, II, and III}
As described earlier, swarm agents move over  a directed graph $\mathcal{G}= (V,E)$. For the purposes of path planning, we define a related graph $\mathcal{G}_s= (V_s,E_s,C_s)$ for which $V_s = V$ and $E_s = E$ and, in addition, the directed edge costs $C_s$ are associated with each edge $e_{ij} \in E_s$. The graph $\mathcal{G}_s$ therefore represents the cost, from the swarm's perspective, of moving from the start to the goal across several edges.
 We consider three types of edge costs and corresponding cost-minimizing paths:
\begin{itemize}
    \item \emph{(Type I): Minimum Accumulated Cost Paths (Weighted Risk and Distance-Based Cost)}:
    Let
    \begin{equation}
    C_s( e_{ij}; \alpha ) := \alpha w_j + (1-\alpha) d_{ij}
\label{eqn:edge_cost}
\end{equation}
denote the edge cost for an edge $e = (i,j) \in E_s$, where $d_{ij} = \texttt{Length}(i,j)/\Delta$ is a dimensionless length.  Thus, the cost of traversing an edge from node $i$ to $j$ is the weighted cost of the distance traveled and the risk incurred arriving at $j$.
    The total cost of a path $\pi$ is therefore
    \begin{equation}
     J_{\rm I}(\pi ; \alpha) := \sum_{e \in \pi} C_s( e ; \alpha )  \;.
     \notag
\end{equation}
Since the cost is a standard accumulation (sum) of costs along the path edges, standard shortest-path planning algorithms can be used to solve for Type I paths.  The parameter $\alpha$ allows trading off emphasis on risk-reduction versus length-reduction in path planning.
    \item \emph{(Type II): Min-Max Risk Paths}:  For Type II paths we set $\alpha = 1$ in  \eqref{eqn:edge_cost} so that the edge cost is purely risk-based. Moreover, we define the cost of an entire path as a maximum rather than a sum:
\begin{equation}
     J_{\rm II}(\pi) := \max_{e \in \pi} C_s( e ; \alpha = 1 )  \;.
     \label{eq:riskII}
\end{equation}
Intuitively, this choice prioritizes the single most dangerous cell and does not consider how long the agent stays in danger across multiple cells. Since it only takes one interaction event to disable a swarm agent this approach is the most risk-averse. However, since  \eqref{eq:riskII} is not a sum along path edges, it is not a standard shortest-path problem. This is known as the bottleneck shortest path in the literature. As described in \cite{pollack1960maximum,duan2009fast}, all bottleneck problems can be solved using shortest path algorithms with an appropriate change of distance function.
\item \emph{(Type III): Minimum Length and Max-Risk-Constrained Paths}: For Type III paths we set $\alpha = 0$ in  \eqref{eqn:edge_cost} so that the edge cost is purely distance-based.
    The total cost to minimize is the length
    \begin{equation}
     J_{\rm III}(\pi ; \alpha = 0) := \sum_{e \in \pi} C_s( e ; \alpha = 0 )  \;,
     \notag
\end{equation}
subject to the constraint that the maximum risk along any node in the path is sufficiently low
\begin{equation*}
J_{\rm II}(\pi) < \Gamma \;.
\end{equation*}
The path planning approach to find a Type III path is to first identify all the nodes (and edges) in $\mathcal{G}_s$ for which $w_i \geq \Gamma$. All of these nodes (and corresponding edges) are removed from the graph to give $\mathcal{G}_l$. By construction, all paths $\pi$ on $\mathcal{G}_l$ satisfy $ J_{\rm II}(\pi) < \Gamma$. Subsequently, a standard shortest-path planning algorithm can be run on $\mathcal{G}_l$ to identify $\pi^*$ that minimizes  $J_{\rm III}(\pi ; \alpha=0)$.
\end{itemize}

\subsubsection{Determining Path Feasibility}
In the stochastic problem setting of this paper, one cannot guarantee that a path which successfully reaches the goal is strictly feasible (i.e., guaranteed to not disable an agent). There is always probability mass of risk region presence, associated with all cells, since the Bayesian updates will always produce strictly positive probabilities, i.e., $g_i > 0$. Thus, feasibility of a path is defined according to a chance constraint.

Feasibility implies the chance of surviving $(D = 0)$ for all nodes along the path. To correctly evaluate this probability, we would need to formulate the joint probability of $\{D_1 =0, D_2=0, \ldots, D_Q=0\}$ for all nodes in a path $\pi$.
In general, $D_i$ and $D_j$ are dependent random variables because they may both be within the neighborhood of a generator at a cell $k$ (i.e., it is possible that both $i,j \in \rho(k)$ in which case the outcomes $D_i$ and $D_j$ are related and depend on the probability of a generator in cell $k$, $g_k$.
Let $\mathcal{N}(\pi)$ denote the set of possible generator cells that could influence an agent following path $\pi = \{u_1, \ldots, u_Q\}$, i.e.,
\begin{equation*}
    \mathcal{N}(\pi) := \bigcup_{k=1}^Q \rho(u_k)\;.
\end{equation*}
For each candidate generator cell $j \in \mathcal{N}(\pi)$ define
\begin{equation*}
   n_j(\pi) := \{ u \in\pi~:~u\in\rho(j) \}
\end{equation*}
as the set of cells along the path that are threatened by a possible generator at $j$ (where $\pi$ is assumed to be a simple path with no duplicate vertices). Let $S_j(\pi)\in \{0, 1\}$ be a binary random variable that denotes the event of cell $j$ not disabling an agent $(S_j =1)$ along path $\pi$ (i.e., any disabling events, if they occur along path $\pi$, do not occur because of the generator at $j$).   If there is no  generator at cell $j$, then
\begin{equation}
    P(S_j=1|G_j = 0)=1\;.
    \label{eq:Psj1Gj0}
\end{equation}
Conversely, if there is a generator at cell $j$ then the probability that it does not disable the agent along a path $\pi$ depends on the number of steps $|n_j(\pi)| > 0$ and is given by
\begin{equation}
    P(S_j=1|G_j = 1)= (1-p_f)^{|n_j(\pi)|}\;.
    \label{eq:Psj1Gj1}
\end{equation}
If $|n_j(\pi)| = 0$ this implies generator $j$ is too far away to affect the agent along path $\pi$ and
\begin{equation*}
    P(S_j=1|G_j = 1)= 1 \;.
\end{equation*}
Furthermore, if there is no generator at cell $j$ then $P(S_j=1|G_j = 0)= 1$ and $P(S_j=0|G_j = 0)= 0$.
Using \eqref{eq:Psj1Gj0} and \eqref{eq:Psj1Gj1} with the law of total probability:
\begin{align}
    P(S_j =1) &= P(S_j=1|G_j = 0) P(G_j=0) + P(S_j=1|G_j = 1) P(G_j=1) \notag \\
    &= 1 (1-g_j) + (1-p_f)^{|n_j(\pi)|} g_j \notag \\
    &= 1-g_j + (1-p_f)^{|n_j(\pi)|} g_j  \;.
    \label{eq:pSj=1}
\end{align}
Now, the feasibility of the path can be stated as the probability that $S_j = 1$ for all $j \in \mathcal{N}(\pi)$ and all of these are independent events, so then using \eqref{eq:pSj=1}:
\begin{equation}
    \mathcal{F}_{\rm exact}(\pi) := \prod_{j \in \mathcal{N}(\pi)} P(S_j=1) =  \prod_{j \in \mathcal{N}(\pi)} \left( 1-g_j + (1-p_f)^{|n_j(\pi)|} g_j  \right) \;
    \label{eq:feasibility_full}
\end{equation}
under the independent cell-wise belief assumption in this paper.
While \eqref{eq:feasibility_full} is the exact calculation required to compute the chance of surviving along path $\pi$, several approximations can be made to obtain lower-cost solutions. For example, if $D_i$ and $D_j$ are assumed independent along the path, then
\begin{equation}
    \mathcal{F}_{\rm approx,I}(\pi)  :=  \prod_{j \in \mathcal{N}(\pi)} \left( 1-p_fg_j  \right)^{|n_j(\pi)|}
    = \prod_{i \in \pi} \prod_{j \in \rho(i)} (1-p_fg_j) \;.
    \label{eq:feasibility_full2}
\end{equation}
The expression \eqref{eq:feasibility_full2} can be proved to be an under-approximation of \eqref{eq:feasibility_full}, i.e., $\mathcal{F}_{\rm approx,I}(\pi) \leq \mathcal{F}_{\rm exact}(\pi)$. To show this fact, let the difference between the two terms within the respective products be:
\begin{align*}
    h(p_f, g_j) &:= \left( 1-p_fg_j  \right)^{|n_j(\pi)|}
    -
    \left( 1-g_j + (1-p_f)^{|n_j(\pi)|} g_j  \right)  \notag \\
     &=
     \left( 1-p_fg_j  \right)^{|n_j(\pi)|}
     -
     \left( 1-g_j \{1 - (1-p_f)^{|n_j(\pi)|} \} \right) \;.
\end{align*}
Since $p_f$ and $g_j$ are both probabilities then $p_f, g_j \in [0,1]$. Suppose $p_f$ is  fixed and consider how the function varies with $g_j$. We find that at the endpoints, $h(p_f, g_j=0) = 0$ and at $h(p_f, g_j=1) = 0$. Thus, at these two boundaries the approximation is equal to the exact value. Now take the derivative of $h$ with respect to $g_j$:
\begin{align*}
    \frac{\partial h}{\partial g_j} =  -p_f|n_j(\pi)|(1- p_f  g_j)^{|n_j(\pi)|-1}+(1-(1-p_f)^{|n_j(\pi)|})
\end{align*}
and the second partial derivative is
\begin{align*}
    \frac{\partial^2 h}{\partial g_j ^2} =
      p_f^2|n_j(\pi)|(|n_j(\pi)|-1)(1- p_f  g_j)^{|n_j(\pi)|-2} \;.
\end{align*}
Suppose that $|n_j(\pi)|=1$. Then, $\partial h/\partial g_j = -p_f + 1 - (1-p_f) = 0$ and ${\partial^2 h}/{\partial g_j ^2} = 0$. This implies that if $|n_j(\pi)|=1$ then $h(p_f,g_j) = 0$ for all $g_j \in[0,1]$. If, instead, $|n_j(\pi)|\geq 2$ then, by inspection,  ${\partial^2 h}/{\partial g_j ^2} \geq 0$. This implies that the quantity $h(p_f, g_j)$ is convex between the two points $g_j = 0$ and $g_j = 1$ for all $p_f \in[0,1]$. If a function is zero at the endpoints and is convex between those endpoints then it must be non-positive. We can conclude that $h(p_f,g_j) \leq 0$ on this interval. This condition relates each term within the products of \eqref{eq:feasibility_full2} and \eqref{eq:feasibility_full} and, as a result, we conclude that $\mathcal{F}_{\rm approx,I}(\pi) \leq \mathcal{F}_{\rm exact}(\pi) $. Moreover, in the special case that $p_f = 1$, and using \eqref{eq:RR_prob} the equation \eqref{eq:feasibility_full2} reduces to
\begin{equation}
    \mathcal{F}_{\rm approx,II}:=\mathcal{F}_{\rm approx,I}(\pi)\bigg|_{p_f = 1}  =  \prod_{i \in \pi} (1-w_i) \;.
    \label{eq:feasibility}
\end{equation}
Since $p_f < 1$ implies that $1-g_j \leq 1 - p_f g_j$, then we can conclude that $\mathcal{F}_{\rm approx,II}(\pi) \leq \mathcal{F}_{\rm approx,I}(\pi)$.

Let $0 < \beta < 1$ be a positive constant. The feasibility can then be treated as a chance constraint--if $\mathcal{F}(\pi) \geq \beta$ we consider the path feasible replacing $\mathcal{F}$ with either  \eqref{eq:feasibility_full},  \eqref{eq:feasibility_full2}, or \eqref{eq:feasibility} as appropriate.

\subsubsection{Phase 1: Algorithm to Identify a Feasible Path}
As mentioned earlier, the first path planning phase aims to identify a feasible path. Our approach is to use a combination of path types I, II, and III with probabilities $q_{\rm I}$, $q_{\rm II}$, and $q_{\rm III}$ respectively (and the constraint that $q_{\rm I} + q_{\rm II} +  q_{\rm III} = 1$). The proposed path planning algorithm is as follows:
\\
\\
{\bf Algorithm:} \texttt{FeasiblePathPlanner}
\begin{enumerate}
\item Initially, for the first  agent ($n=1$),  define a fixed probability $0 < p_0 < 1$ for all generators (i.e., $g_j \gets p_0$ for all $j \in V$). For example, if $\hat M$ generators are estimated in the area, a uniform probability of $p_0 = \hat M/S^2$ may be assigned. Alternatively, other sources of prior information can be incorporated in defining this prior.
\item Use the prior probability to compute the risk-region probabilities, $w_i$ for $i \in V$ from \eqref{eq:RR_prob}.
\item Plan a path $\hat \pi_n$
\begin{enumerate}[label=\alph*)]
\item With probability $q_{\rm I}$ plan a Type I path with parameter $\alpha$.
\item With probability $q_{\rm II}$ plan a Type II path.
\item With probability $q_{\rm III}$ plan a Type III path. To do so, first plan a Type II path $\hat{\pi}_{\rm II}$ and record the cost $J_{\rm II}(\hat{\pi}_{\rm II})$. Then set $\Gamma = J_{\rm II}(\hat{\pi}_{\rm II}) + \eta$ and plan a Type III path where $\eta > 0$ is a risk buffer.
\end{enumerate}
\item Execute the path to obtain $\pi_n \subseteq \hat \pi_n$. Let $V(\pi_n) = \{u_1, \ldots, u_{Q}\}$ denote the set of nodes sequentially visited along the path.
\item For each node $i \in \{u_1, \ldots, u_{Q-1}\}$
compute $\rho(i)$ and apply the update \eqref{eq:prob_gj=1|di=0} to obtain the new probabilities $g_j'$ for all $j \in \rho(i)$.
\item For the node $u_Q$, if $u_Q$ is not the goal node then compute $\rho(u_Q)$ and apply the update \eqref{eq:prob_gj=1|di=1} to obtain the new probabilities $g_j'$ for all $j \in \rho(u_Q)$. Otherwise, if $u_Q$ is the goal node then compute $\rho(u_Q)$ and apply the update \eqref{eq:prob_gj=1|di=0} to obtain the new probabilities $g_j'$ for all $j \in \rho(u_Q)$. Note that when computing \eqref{eq:prob_gj=1|di=1} for each cell $j \in \rho(u_Q)$ the pre-update values $g_j$ are used (rather than the sequentially updated $g_j'$ values).
\item With the new generator probabilities, update the risk-region probabilities $w_i$ for $i \in V$ from \eqref{eq:RR_prob}.
\item Update the swarm planning graph $\mathcal{G}_{s}$ with weights to reflect the new $w_i$.
\item If the goal node was reached along $\pi_n$ then calculate $\mathcal{F}(\pi_n)$ using \eqref{eq:feasibility_full} (or approximate via \eqref{eq:feasibility_full2} or \eqref{eq:feasibility}). If $\mathcal{F}(\pi_n) \geq \beta$, then return the feasible path $\hat \pi_n$.
\item If $n = n_{{\rm max},1}$ then return \texttt{Infeasible}. Otherwise, increment $n \gets n + 1$ and return to Step 3.
\end{enumerate}

The above algorithm is controlled via the parameters: $\alpha$, $\beta,$ $\eta$, $n_{{\rm max},1}$, $q_{\rm I}$, $q_{\rm II}$, and $q_{\rm III}$. The choices of the probabilities $q_{\rm I}$, $q_{\rm II}$, and $q_{\rm III}$ control  the degree to which path planning diversifies the types of paths  in Step 3. A random number generator is used with threshold values of $q_{\rm I}$, $q_{\rm II}$, and $q_{\rm III}$ to determine which type of path to plan at each iteration. The parameter $n_{{\rm max},1}$ determines the maximum number of iterations (equivalent agents deployed) before the algorithm terminates with an infeasible solution returned.
The parameter $\alpha$ controls the priority of path length versus risk, and the parameter $\beta$ controls the degree of confidence desired when ensuring a path to the goal is actually feasible.  The parameter $\eta$ slightly increases the risk $\Gamma$ from the bottleneck solution so that the pruned graph (under the $J_{\rm II}(\pi) < \Gamma$ condition) remains connected and Type III planning can find shorter paths with nearly minimal maximum risk.

\subsubsection{Phase 2: Algorithm to Reduce the Length of a Feasible Path}
Additionally, once a feasible path is found we can further reduce the length of the path by iteratively running the Type III path planner. In effect, additional agents are used to explore nearby trajectories and gain possible length savings. Suppose that $\hat \pi^*_0$ is a known feasible path (satisfying $\mathcal{F}(\hat \pi^*_0) \geq \beta$) and the generator and risk region probabilities, $g_i$ and $w_i$, are known for all cells $i \in \{1, \ldots, S^2\}$ (i.e., these quantities are the outcome of the \texttt{FeasiblePathPlanner}). By  re-computing $g_i, w_i$ after each path the agents can sequentially explore path-reductions with an acceptable level of risk by solving for Type III paths.
The proposed path planning algorithm is as follows where we refer to $\hat \pi^*_0$ as $\hat \pi_0$:
\\
\\
{\bf Algorithm:} \texttt{RiskAwarePathLengthReduction}
\begin{enumerate}
\item At the first instance set $n = 1$, $\gamma  = J_{\rm II}( \hat \pi_0)$, $\hat \pi_{\rm best} \gets \hat \pi_0$, and $L_{\rm best} \gets \texttt{Length}( \hat \pi_0)$.
\item Set the risk threshold as $\Gamma \gets {\rm min}(\gamma + \epsilon,1)$, where $\epsilon \in (0, \beta)$ is a user-specified parameter.
\item Identify all the cells where the risk exceeds the threshold, i.e., $\mathcal{V} = \{ i \in V~:~w_i \geq \Gamma\}$ and the corresponding edges $\mathcal{E} = \{ (i,j) \in E~:~ i \in \mathcal{V} \text{ or } j \in \mathcal{V} \}$.
\item Create the modified graph $\mathcal{G}_l = (V_l, E_l)$ where $V_l = V \setminus \mathcal{V}$ and $E_l = E \setminus \mathcal{E}$.
\item Compute a new shortest length path in $\mathcal{G}_l$ that joins the start and the goal. The resulting path $\hat \pi_n$ minimizes the total length (i.e., using distance as a cost). By construction, the new path is guaranteed to satisfy the risk threshold, i.e., $J_{\rm II}(\hat \pi_n) < \Gamma$.
\item Deploy a swarm agent to traverse the  path to obtain $\pi_n \subseteq \hat \pi_n$.
\item Using the data $\pi_n$, perform an update to obtain $g_i$ and $w_i$ (via Steps 5--7 in \texttt{FeasiblePathPlanner}).
\item If the swarm agent successfully reached the goal along $\pi_n$ then:
\begin{enumerate}
    \item Set $L_n$ as the length of the path $\hat \pi_n$.
    \item Calculate $\mathcal{F}(\hat \pi_n)$ using \eqref{eq:feasibility_full} (or approximate via \eqref{eq:feasibility_full2} or \eqref{eq:feasibility}).
    \item If $\mathcal{F}( \hat \pi_n) \geq  \beta$ and if $L_n < L_{\rm best}$, then set $\Delta L \gets L_{\rm best} - L_n$, $L_{\rm best} \gets L_n$, $\hat \pi_{\rm best} \gets \hat \pi_n$, and $\gamma  \gets J_{\rm II}(\hat \pi_{\rm best})$.  Otherwise, set $\Delta L \gets 0$.
    \item If $\mathcal{F}( \hat \pi_n) \geq \beta$ and if $\Delta L < \delta$ then return $\hat \pi_{\rm best}$.
\end{enumerate}
    \item If $n = n_{{\rm max},2}$ then return $\hat \pi_{\rm best}$.
\item Increment $n \gets n  + 1$, and return to Step 2.
\end{enumerate}
The above algorithm is controlled via the parameters $\epsilon$, $n_{{\rm max},2}$, and $\delta$. The parameter $\epsilon$ determines the risk-taking tolerance of subsequent agents. Higher values will prompt agents to explore riskier regions in search of a shorter path and may lead to faster convergence but possibly more attrition. The parameter $n_{{\rm max},2}$ is a maximum on the number of agents to be expended in search of path reductions. Alternatively, the parameter $\delta$ can be used to define a distance threshold where the path is determined to be converged. In certain edge cases, it is possible that Steps 3 and 4 above effectively disconnect the start and goal nodes---in such cases subsequent path planning will return infeasible and we assume that no further path reduction is possible (the solver returns $\hat \pi_{\rm best}$).
\section{Results}
\label{sec:results}
This section presents an illustrative example of the path planner on the scenario shown in Fig.~\ref{fig:setup}. This example takes place over a $64 \times 64$ grid environment with $\hat M = 8$ circular risk regions of radius $R = 10$ length units (LU) and cell spacing $\Delta = 1~\text{LU}$. The  black circles in Fig.~\ref{fig:setup} denote the boundaries of the risk regions and the magenta markers are the generator points. The risk-free shortest path (computed with perfect knowledge of the environment) is shown in red in Fig.~\ref{fig:setup} and has a length $L = 147.15~\text{LU}$. Because the path length is computed as the sum of edge costs over the discretized environment, there are multiple paths with the same minimum length. Additional simulation parameters are listed in Table~\ref{tab:scitech_params}. The generator prior $p_0$ is computed by assuming there are $M=8$ risk regions and dividing by the total number of cells $64^2 = 4096$. The chance constraints in this example were calculated using \eqref{eq:feasibility}.
    \begin{table}[H]
    \centering
    \caption{Simulation and solver parameters for the illustrative example.}
    \label{tab:scitech_params}
    \begin{tabular}{l  c  l}
    \hline\hline
    {\bf Parameter} & {\bf Symbol} & {\bf Value} \\
    \hline
    Risk-region centers $x$ position & $x_m$ & $\{15.5, 30.5, 40.5, 20.5, 38.5, 62.5, 62.5,
    33.5\}$ \\
    Risk-region centers $y$ position & $y_m$ & $\{35.5, 7.5, 50.5, 20.5, 35.5, 20.5, 50.5, 62.5\}$ \\
    Disablement probability & $p_f$ & $0.95$ \\
    Uniform generator prior & $p_0$ & $M / S^2 = 8 / 4096$ \\
    Path-type weights & $(q_{\mathrm{I}}, q_{\mathrm{II}}, q_{\mathrm{III}})$ & $(0.40, 0.40, 0.20)$  \\
    Type I risk--distance weight & $\alpha$ & $0.90$ \\
    Type III Phase~1 risk buffer & $\eta$ & $0.01$ \\
    Feasibility parameter & $\beta$ & $0.95$ \\
    Phase~1 max. iterations & $n_{\max,1}$ & $50$ \\
    Phase~2 risk buffer & $\epsilon$ & $0.50$
    \\
    Phase~2 length tolerance & $\delta$ & $0.50~\text{LU}$ \\
    Phase~2 max. iterations  & $n_{{\rm max},2}$ & $10$ \\
    \hline\hline
    \end{tabular}
    \end{table}

\begin{figure}[H]
    \centering
    \begin{subfigure}[b]{0.45\textwidth}
      \centering
      \includegraphics[width=\linewidth]{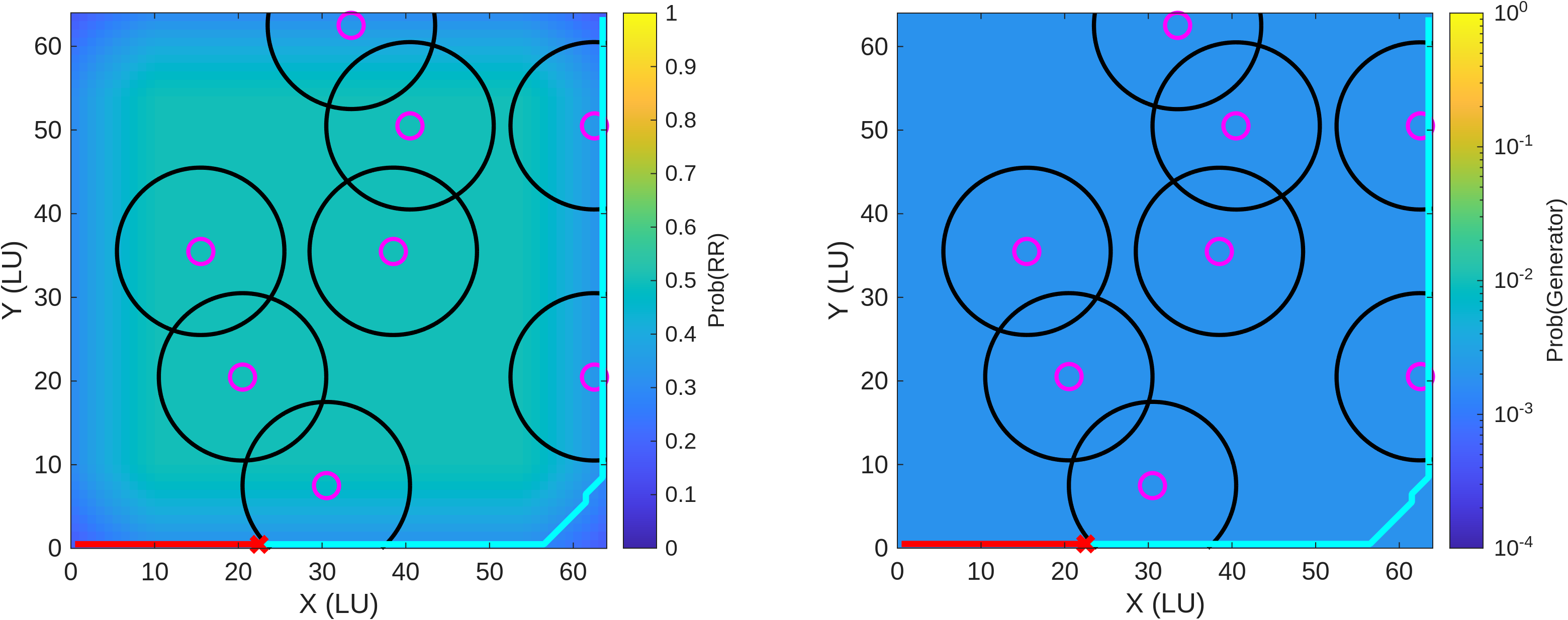}
      \subcaption{Iteration \#1 with 1 agent lost.}
    \end{subfigure}\hspace{8pt}%
    \begin{subfigure}[b]{0.45\textwidth}
      \centering
      \includegraphics[width=\linewidth]{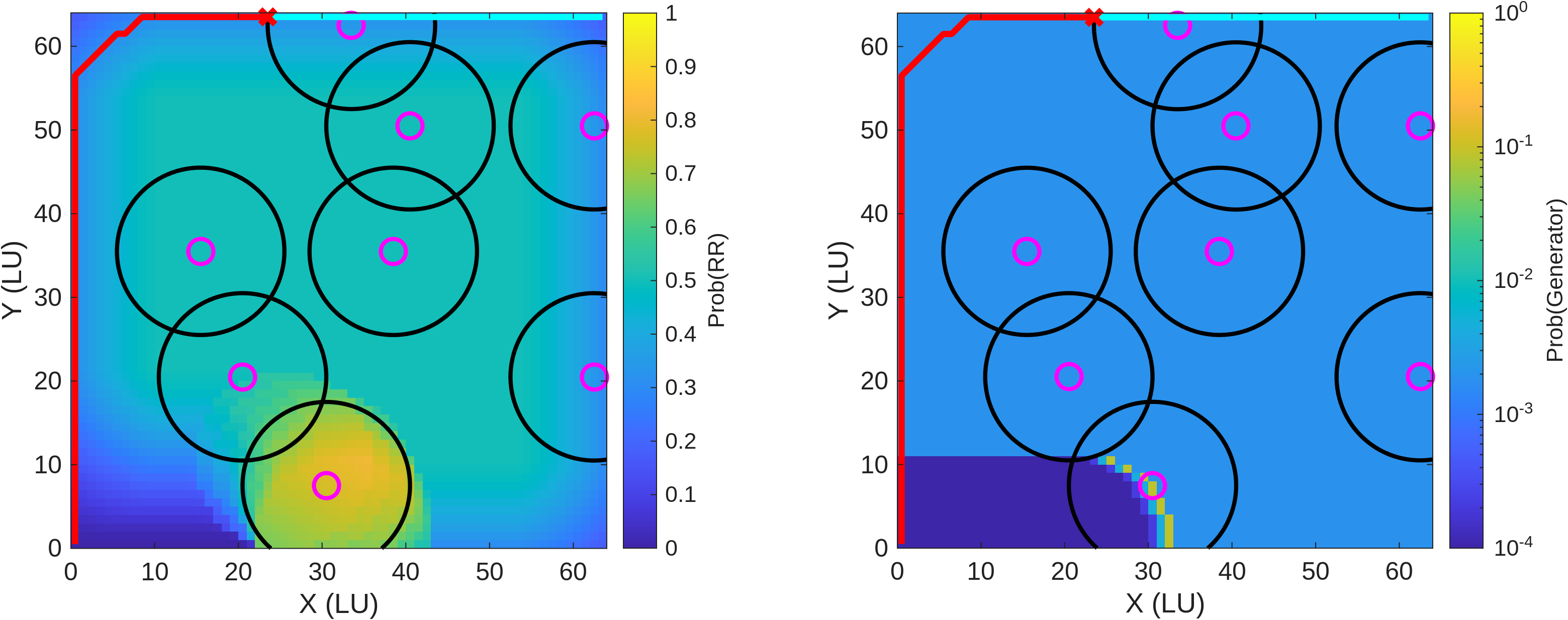}
      \subcaption{Iteration \#2 with 2 agents lost.}
    \end{subfigure}

    \vspace{4pt}

    \begin{subfigure}[b]{0.45\textwidth}
      \centering
      \includegraphics[width=\linewidth]{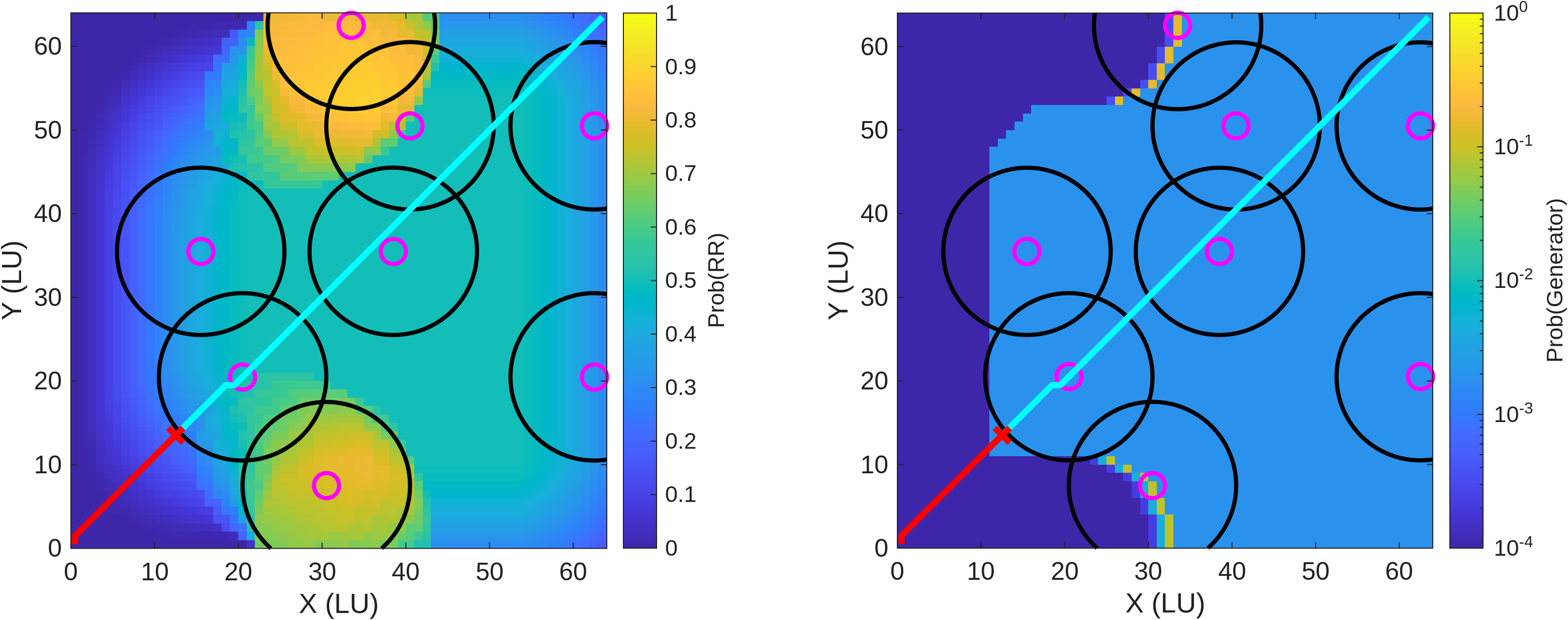}
      \subcaption{Iteration \#4 with 4 agents lost.}
    \end{subfigure}\hspace{8pt}%
    \begin{subfigure}[b]{0.45\textwidth}
      \centering
      \includegraphics[width=\linewidth]{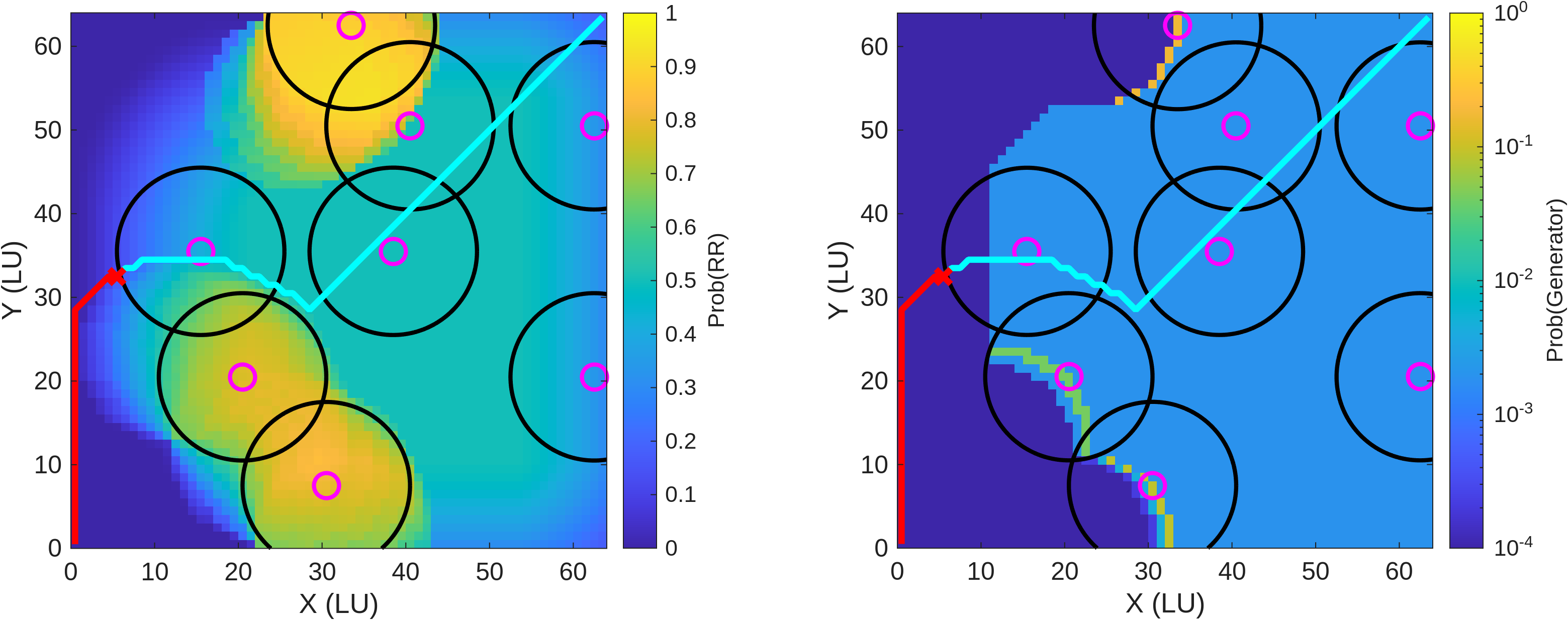}
      \subcaption{Iteration \#6 with 6 agents lost.}
    \end{subfigure}

    \vspace{4pt}

    \begin{subfigure}[b]{0.45\textwidth}
      \centering
      \includegraphics[width=\linewidth]{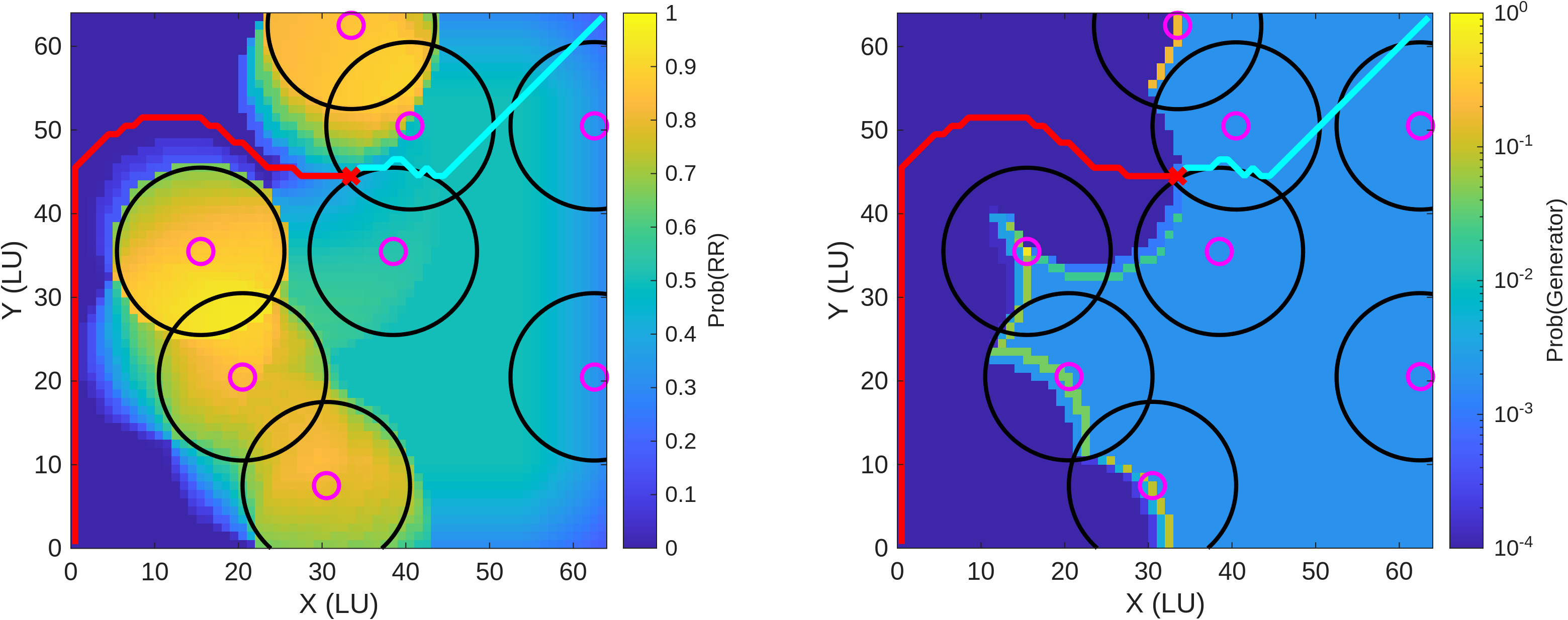}
      \subcaption{Iteration \#8 with 8 agents lost.}
    \end{subfigure}\hspace{8pt}%
    \begin{subfigure}[b]{0.45\textwidth}
      \centering
      \includegraphics[width=\linewidth]{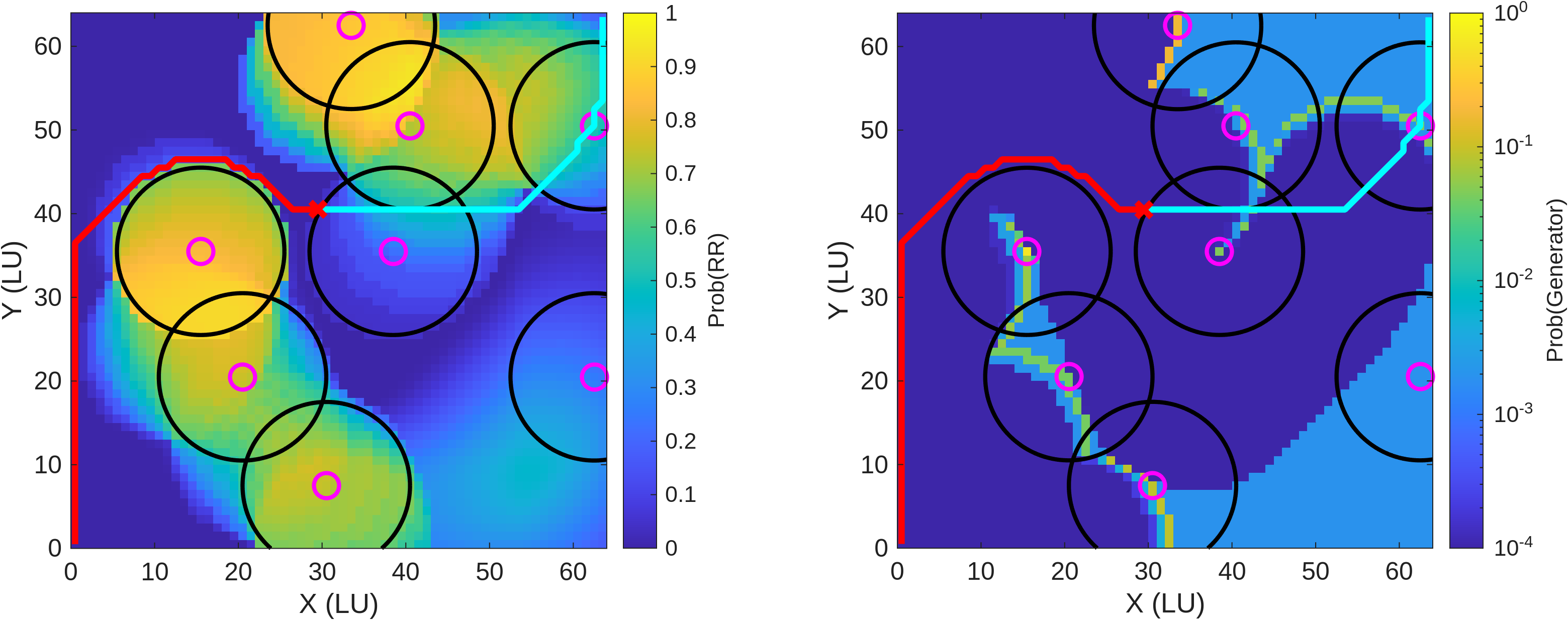}
      \subcaption{Iteration \#10 with 10 agents lost.}
    \end{subfigure}

    \vspace{4pt}

    \begin{subfigure}[b]{0.45\textwidth}
      \centering
      \includegraphics[width=\linewidth]{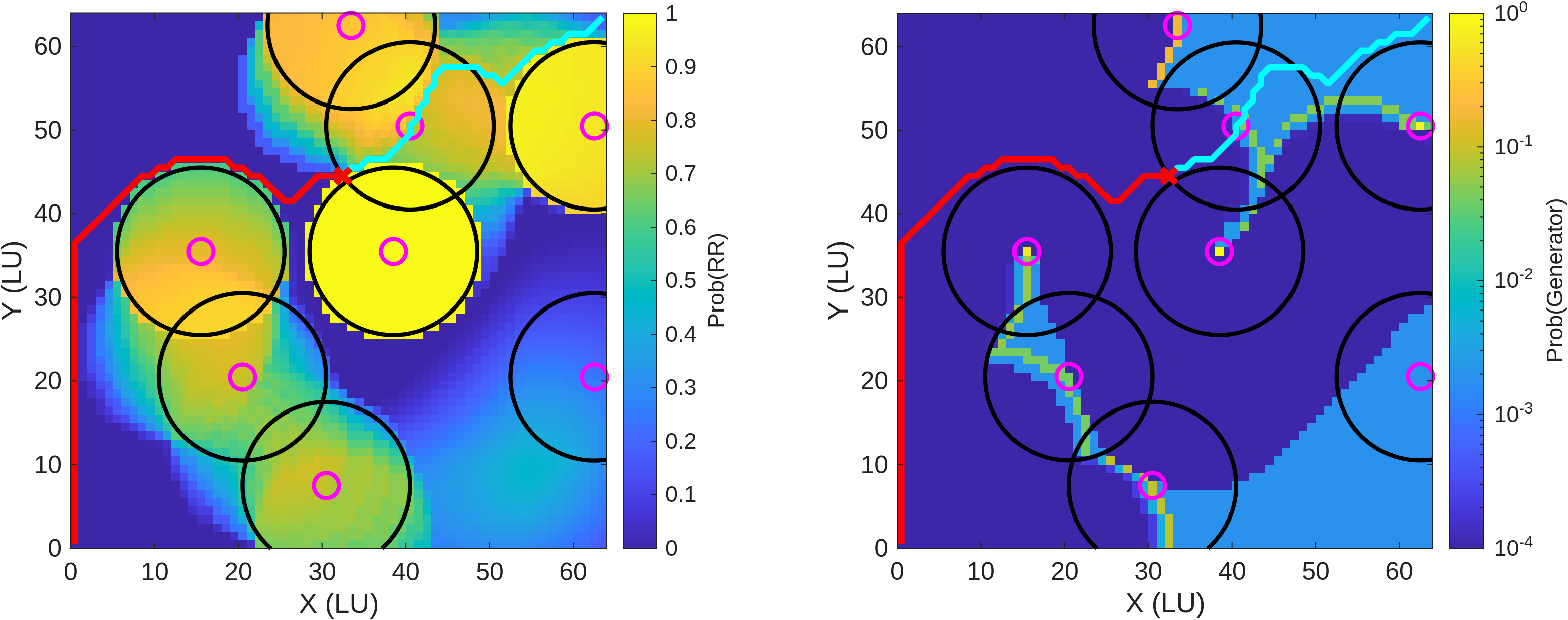}
      \subcaption{Iteration \#12 with 12 agents lost.}
    \end{subfigure}\hspace{8pt}%
    \begin{subfigure}[b]{0.45\textwidth}
      \centering
      \includegraphics[width=\linewidth]{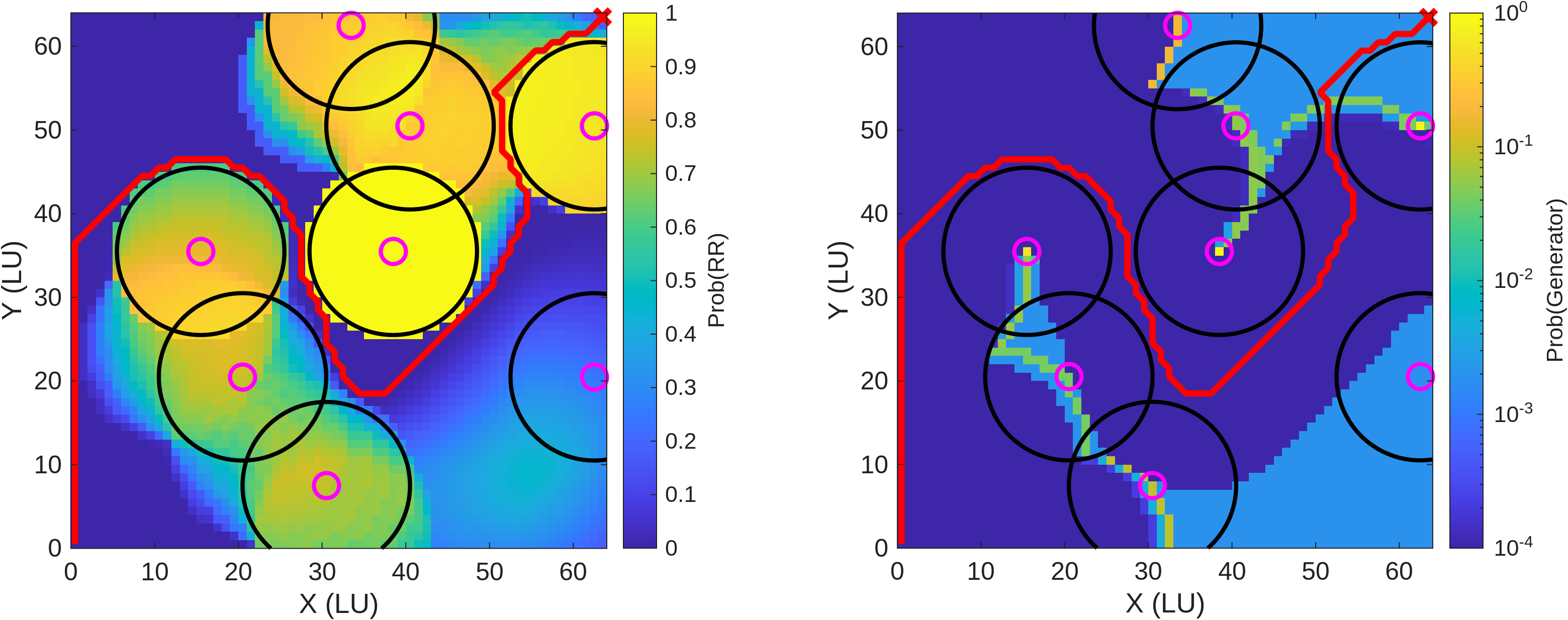}
      \subcaption{Iteration \#13 with 12 agents lost.}
    \end{subfigure}

    \caption{Example Phase~1 simulation (finding a feasible path).}
    \label{fig:filmstrip1}
    \end{figure}

\begin{figure}[H]
    \centering
    \begin{subfigure}[b]{0.45\textwidth}
      \centering
      \includegraphics[width=3in]{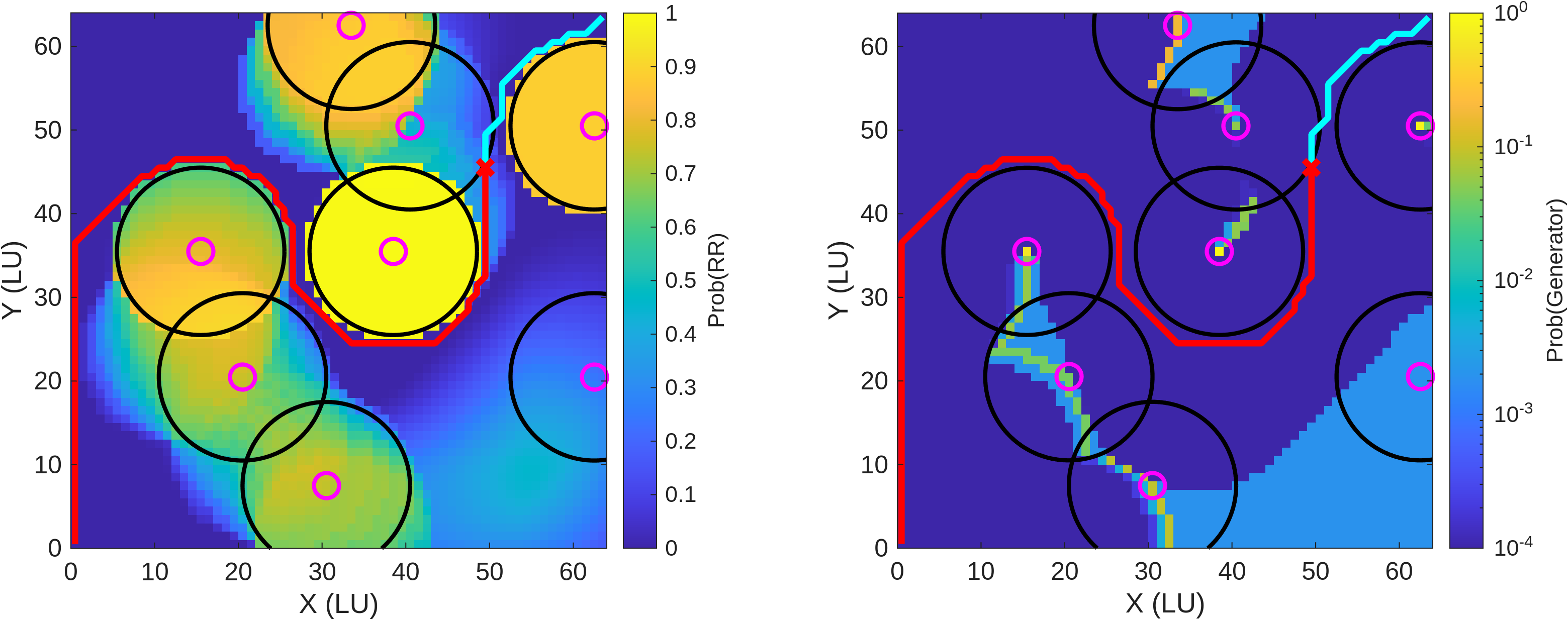}
      \subcaption{Iteration \#14 with 1 agent lost ($L_{\mathrm{best}} = 158.37\text{
      LU}$).}
    \end{subfigure}\hspace{8pt}%
    \begin{subfigure}[b]{0.45\textwidth}
      \centering
      \includegraphics[width=3in]{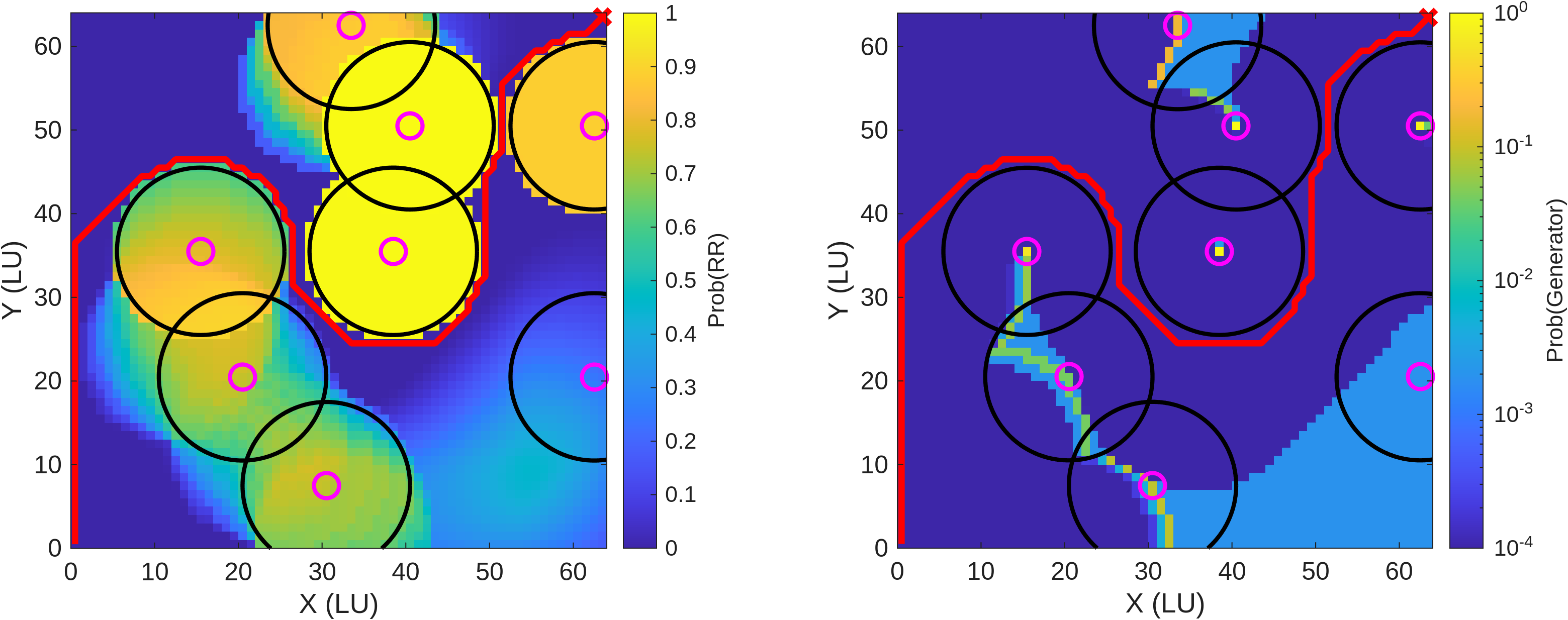}
      \subcaption{Iteration \#15 with goal reached ($L_{\mathrm{best}} = 147.15\text{
      LU}$).}
    \end{subfigure}

    \vspace{6pt}

    \begin{subfigure}[b]{0.65\textwidth}
      \centering
      \includegraphics[width=3in]{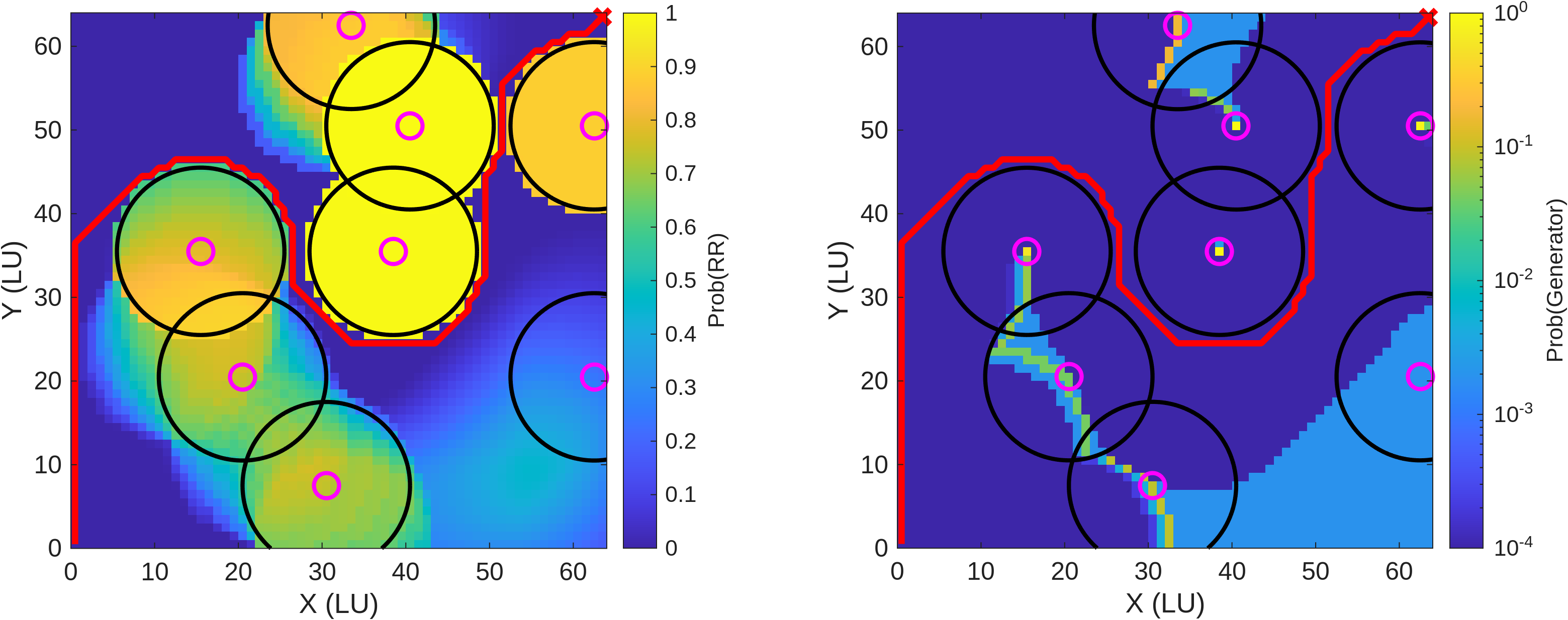}
      \subcaption{Iteration \#16 with goal reached again ($L_{\mathrm{best}} = 147.15\text{
      LU}$).}
    \end{subfigure}

    \caption{Example Phase~2 simulation (reducing the length of a feasible path).}
    \label{fig:filmstrip2}
    \end{figure}

Results are shown in  Fig.~\ref{fig:filmstrip1} for Phase 1 and in  Fig.~\ref{fig:filmstrip2} for Phase 2. Each of these figures shows a snapshot in time of one agent's deployment and two corresponding panels within each sub-figure. The left panel illustrates a colormap of the swarm's understanding of where the risk regions are, i.e., $w_i$ for each cell. The right panel shows the swarm's understanding of where the generators are located, i.e., $g_i$ for each cell with log-scale colormap. Both frames have the true risk-region generator locations shown as magenta circles and the true risk-affected regions shown as black circles, similar to Fig.~\ref{fig:setup}.  In each panel, the cyan path is the path planned from the start to the goal and the superimposed red path is the actual path traversed by the swarm agent, ending at the location where the swarm agent was disabled.

Focusing first on Fig.~\ref{fig:filmstrip1}, we see that in Fig.~\ref{fig:filmstrip1}a the first agent planned to travel along the lower boundary of the space, cut across the lower right corner, and then travel upwards to the goal. The agent plans this path because the boundary of the space has reduced risk. This is a result of a uniform $g_i$ distribution---at the edges the space outside the domain cannot contribute to elevated risk, hence under the uniform generator assumption the boundaries of the domain are the safest given the prior knowledge. However, we see that the agent is lost upon encountering the risk region centered on $x_2 = 30.5$ and $y_2 = 7.5$. Moving over to Fig.~\ref{fig:filmstrip1}b, we now see the new prior map of $w_i$ and $g_i$ that the next agent uses for planning. The region around the first agent's path has reduced generator probability (and hence reduced risk region probability). The disablement point draws an arc of high generator probability that passes through the actual generator center. If the $p_f$ probability was lower, one would expect this arc to be smeared across multiple pixels since the range of the generator would be more uncertain. In Fig.~\ref{fig:filmstrip1}c the swarm now has high probability of two risk regions (bright yellow spots) and the agent decides to traverse the center of the graph with lower risk. This process continues until the 13th iteration where the agent successfully reaches the goal. The evaluation of the posterior in this case shows that the chance constraints are satisfied; however this is not always the case. In some instances multiple agents must reach the goal before the chance constraints are satisfied. The length of the feasible path found at the conclusion of Phase 1 is 158.37~LU.

The simulation in Fig.~\ref{fig:filmstrip2} shows the results of Phase 2, using the \texttt{RiskAwarePathLengthReduction} algorithm to reduce path length. Iteration \#14 in Fig.~\ref{fig:filmstrip2}a shows that the algorithm produces a shorter path but it enters a poorly estimated risk region near the end of the path. The agent is lost following this shorter (but infeasible) path and subsequently the generator and risk maps are updated. On the second iteration the planner correctly routes the agent through the narrow gap near the goal and the agent reaches the goal. The resulting path is now 147.15~LU. However, despite reaching the goal again the convergence criterion is not met. On Iteration \#16 the agent again reaches the goal with the same path length and Phase 2 concludes. The path has been reduced from 158.37~LU to 147.15~LU (a 7.1\% reduction at the cost of one additional agent).

\begin{figure}[H]
    \centering
    \includegraphics[width=0.4\textwidth]{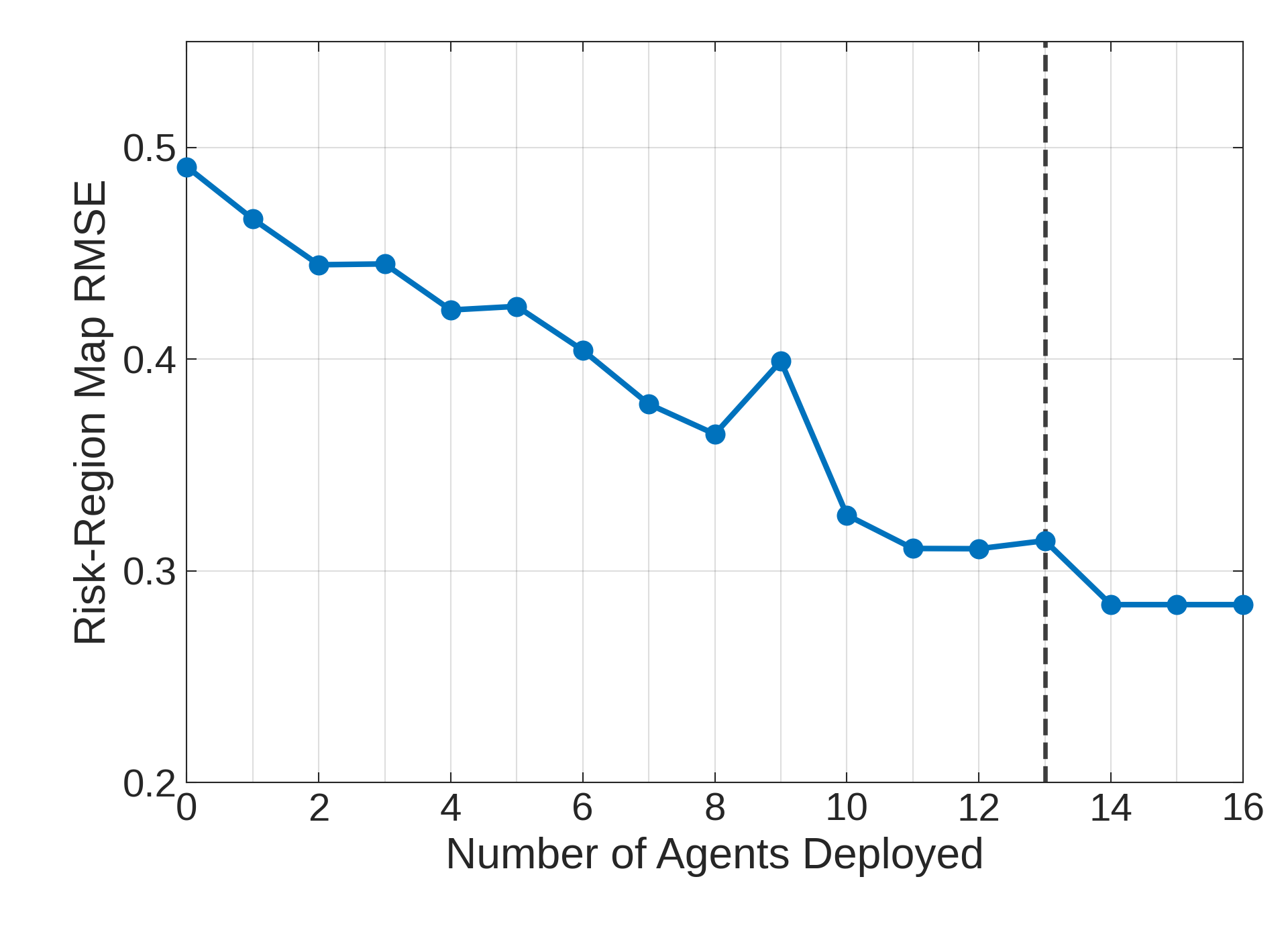}
    \caption{Evolution of the risk-region probability map root-mean-square error (RMSE). The vertical dashed line at agent 13 marks the transition from
    Phase~1 to Phase~2.}
    \label{fig:map_accuracy_evolution}
    \end{figure}

In summary, after completing Phase 1 and Phase 2 there were a total of 16 agents deployed, of which three reached the goal and 13 did not. Of these 13 disabled agents, 12 were lost finding a feasible path and one was lost during the path reduction phase. The path found at the conclusion of Phase 2 has a length equal to the optimal path found with perfect prior knowledge, as shown in Fig.~\ref{fig:setup}. Considering the final panel, Fig.~\ref{fig:filmstrip2}c, we see that the risk region is well defined, with some generators appearing as bright pixels in the generator map (i.e., these are well-localized) and others still having uncertainty along arcs. The lower right risk region/generator was never encountered and thus never localized. Define the following metric for risk-region map root-mean-square error (RMSE):
\begin{equation*}
\mathrm{RMSE} =
\sqrt{\frac{1}{S^2}\sum_{i=1}^{S^2}\left(w_i - O_i\right)^2} \,,
\label{eq:rr_rmse}
\end{equation*}
where $w_i$ is the probability of a risk region in cell $i$ and $O_i$ is the true occupancy of cell $i$ (equal to one if occupied and equal to zero otherwise).
Figure~\ref{fig:map_accuracy_evolution} illustrates the evolution of the map accuracy according to this metric. In most cases, as additional agents are deployed the RMSE reduces. In Phase 2 when the last two agents follow the same path they do not make a significant change to the RMSE.

\section{Conclusion}
\label{sec:conclusion}
This paper investigated algorithms to sequentially deploy agents to identify a path in an environment with unknown risk regions. The path of the agents and the location where they are disabled (prior to reaching the goal) are used in a Bayesian framework to update an evolving map of the risk-region probability over the space. Three different path types were proposed to exploit this risk map either by emphasizing shortest paths, low-risk paths, or combinations thereof. Algorithms were developed to: (a) find a feasible path, and (b) reduce the length of a feasible path once it is found. The illustrative example tested showed that a swarm of 13 agents was sufficient to find a feasible path in the presence of eight risk regions in the particular configuration chosen for the example. Subsequently, a 7.1\% path-length  reduction could be achieved to identify the true optimal path using three more agents of which one was lost.

Future work may relax the assumption that agents are deployed sequentially, and instead allow multiple swarm agents to be deployed simultaneously or without waiting for each previous agent to complete its path. In this context, it may be of interest to optimize
the total time to reach the goal  as the elapsed time since the first swarm agent initiated its path.

\section*{Acknowledgments}
This work was supported by the Air Force Research Laboratory through a subcontract with DCS Corporation. During the preparation of this work, the authors used Antigravity with Gemini  (Google, 2026) to assist with refining an early draft of the simulation code into its current form. All code was subsequently reviewed and verified by the authors.  The AI tool was also used to evaluate author-written drafts of this manuscript by generating  critical review reports that identified copyediting suggestions and potential technical errors. The authors then reviewed these suggestions and used their judgment and analysis to refine the text themselves. The authors take full responsibility for the content of this publication.
\bibliography{refs}

\end{document}